\documentclass[final,5p,times]{elsarticle}

\usepackage{graphicx}
\usepackage{verbatim}

\usepackage{amsmath,amsthm,amssymb}
\usepackage{bbm}
\usepackage{mathtools}
\usepackage{color}
\usepackage[ruled,vlined,linesnumbered]{algorithm2e}
\usepackage[capitalise]{cleveref}
\usepackage{multirow}
\usepackage{balance}
\usepackage{picins}

\RequirePackage{ifmtarg}

\usepackage{url}

\biboptions{sort&compress,square,numbers,comma}

\newcommand{\nth}[2]{\ensuremath{{#1}^{\mbox{\scriptsize #2}}}} 
\newcommand{\ie}{\emph{i.e.}}
\newcommand{\eg}{\emph{e.g.}}

\newcommand{\cf}{\emph{cf.}}
\newcommand{\diff}[2]{\frac{\partial #1}{\partial #2}}
\newcommand{\vctg}[1]{\ensuremath{\boldsymbol{#1}}} 
\newcommand{\vct}[1]{\ensuremath{\mathbf{#1}}}

\newcommand{\set}[1]{\ensuremath{\mathcal{#1}}}

\newcommand{\T}{\ensuremath{\top}}

\newcommand{\ind}[1]{\ensuremath{\mathbbm I [#1]}}
\newcommand{\argmax}{\operatornamewithlimits{\arg\,\max}}

\newcommand{\argmin}{\operatornamewithlimits{\arg\,\min}}

\newcommand{\reals}{\mathbb{R}}
\newcommand{\tsum}{\textstyle\sum}

\makeatletter
\newcommand{\squareSet}[2][]{\@ifmtarg{#1}{\ensuremath{\left[#2\right]}}
  {\ensuremath{\left[\vphantom{#1}#2 \; \right|\left.\,#1\vphantom{#2}\right]}}
}
\makeatother
\newcommand{\funcName}[1]{\ensuremath{#1}}
\makeatletter
\newcommand{\funcArg}[1]{\@ifmtarg{#1}{\cdot}{#1}}
\makeatother
\newcommand{\functionTwo}[3]{\ensuremath{#1\left(\funcArg{#2},\funcArg{#3}\right)}}
\newcommand{\functionOne}[2]{\ensuremath{#1\left(\funcArg{#2}\right)}}

\newcommand{\expect}[3][]{\mathbb{E}_{#2}\squareSet[#1]{\funcArg{#3}}}
\newcommand{\lossChr}{\funcName{\ell}}
\newcommand{\loss}[3][]{\functionTwo{\lossChr_{#1}}{#2}{#3}}
\newcommand{\riskChr}{\funcName{R}}
\newcommand{\risk}[2]{\functionTwo{\riskChr}{#1}{#2}}
\newcommand{\emprisk}[2]{\functionTwo{\hat{\riskChr}}{#1}{#2}}
\newcommand{\regularizer}[1]{\functionOne{\Omega}{#1}}
\newcommand{\objChr}[1][\learnAlgo]{\funcName{V}_{#1}}
\newcommand{\regObj}[3][\learnAlgo]{\functionTwo{\objChr[#1]}{#2}{#3}}

\newcommand{\hyp}{\ensuremath{f}}
\newcommand{\learnAlgo}{\ensuremath{\mathfrak{L}}}
\newcommand{\hypSpace}{\ensuremath{\mathcal{F}}}

\newcommand{\trainDS}{\set D_{\textrm{tr}}}
\newcommand{\valDS}{\set D_{\textrm{vd}}}
\newcommand{\regParam}{C}

\journal{Neurocomputing, {\rm \url{https://doi.org/10.1016/j.neucom.2014.08.081}}}

\begin{document}

\begin{frontmatter}



\title{Support Vector Machines under Adversarial Label Contamination}
\author[tum]{Huang Xiao}\ead{xiaohu@in.tum.de}
\author[ca]{Battista Biggio\corref{cor1}}\ead{battista.biggio@diee.unica.it}
\author[ca]{Blaine Nelson}\ead{blaine.nelson@gmail.com}
\author[tum]{Han Xiao}\ead{xiaoh@in.tum.de}
\author[tum]{Claudia Eckert}\ead{claudia.eckert@in.tum.de}
\author[ca]{Fabio Roli}\ead{roli@diee.unica.it}
\address[tum]{Department of Computer Science, Technical University of Munich, Boltzmannstr. 3, 85748, Garching, Germany}
\address[ca]{Department of Electrical and Electronic Engineering, University of Cagliari, Piazza d'Armi, 09123, Cagliari, Italy}

\cortext[cor1]{Corresponding author}

\begin{abstract} 
  Machine learning algorithms are increasingly being applied in
  security-related tasks such as spam and malware detection,
  although their security properties against deliberate attacks have
  not yet been widely understood.  Intelligent and adaptive attackers
  may indeed exploit specific vulnerabilities exposed by machine
  learning techniques to violate system security. Being robust to
  adversarial data manipulation is thus an important, additional
  requirement for machine learning algorithms to successfully operate
  in adversarial settings.  In this work, we evaluate the security of
  Support Vector Machines (SVMs) to well-crafted, adversarial label
  noise attacks. In particular, we consider an attacker that aims to
  maximize the SVM's classification error by flipping a number of
  labels in the training data. We formalize a corresponding optimal
  attack strategy, and solve it by means of heuristic approaches to
  keep the computational complexity tractable.  We report an extensive experimental analysis on the
  effectiveness of the considered attacks against linear and non-linear SVMs, both on synthetic and real-world
  datasets. We finally argue that our approach can also provide useful insights for developing more secure SVM learning algorithms, and also novel techniques in a number of related research areas, such as semi-supervised and active learning.
\end{abstract}


\end{frontmatter}

\section{Introduction}
\label{sect:introduction}
 
Machine learning and pattern recognition techniques are increasingly
being adopted in security applications like spam, intrusion and
malware detection, despite their security to adversarial attacks has
not yet been deeply understood. In adversarial settings, indeed,
intelligent and adaptive attackers may carefully target the machine
learning components of a system to compromise its security.  Several
distinct attack scenarios have been considered in a recent field of
study, known as \emph{adversarial machine
  learning}~\cite{barreno-ASIACCS06,barreno10,huang11,biggio14-tkde}.
For instance, it has been shown that it is possible to gradually
\emph{poison} a spam filter, an intrusion detection system, and even a
biometric verification system (in general, a classification algorithm)
by exploiting update mechanisms that enable the adversary to
manipulate some of the training
data~\cite{nelson08,nelson09,rubinstein09,biggio12-spr,biggio13-icb,kloft10,kloft12b,biggio12-icml,biggio14-svm-chapter};
and that the detection of malicious samples by linear and even some
classes of non-linear classifiers can be \emph{evaded} with few
targeted manipulations that reflect a proper change in their feature
values \cite{biggio13-ecml,biggio14-svm-chapter,dalvi04,lowd05,nelson12-jmlr}.
Recently, poisoning and evasion attacks against clustering algorithms have also been formalized to show that malware clustering approaches can be significantly vulnerable to well-crafted attacks~\cite{biggio13-aisec,biggio14-spr}.

Research in adversarial learning not only investigates the security
properties of learning algorithms against well-crafted
attacks, but it also focuses on the development of more secure learning
algorithms.
For evasion attacks, this has been mainly achieved by explicitly
embedding knowledge into the learning algorithm of the possible data
manipulation that can be performed by the attacker, \eg, using
game-theoretical models for
classification~\cite{dalvi04,globerson-ICML06,teo08,bruckner12},
probabilistic models of the data distribution drift under
attack~\cite{biggio11-smc,rodrigues09}, 
and even multiple classifier systems~\cite{kolcz09,biggio10-mcs,biggio-IJMLC10}.
Poisoning attacks and manipulation of the training data have been differently countered with data sanitization (\ie, a form of outlier
detection)~\cite{cretu08,nelson08,nelson09}, multiple classifier
systems~\cite{biggio11-mcs}, and robust
statistics~\cite{rubinstein09}.
Robust statistics have also been exploited to formally show that the  \emph{influence function} of SVM-like algorithms can be bounded under certain conditions~\cite{christmann04}; \eg, if the kernel is bounded. This ensures some degree of robustness against small perturbations of training data, and it may be thus desirable also to improve the security of learning algorithms against poisoning.

In this work, we investigate the vulnerability of SVMs to a specific
kind of training data manipulation, \ie, worst-case label noise. 
This can be regarded as a carefully-crafted attack in which the labels of a subset of the training data are flipped to maximize the SVM's classification error.  While stochastic label
noise has been widely studied in the machine learning literature, to
account for different kinds of potential labeling errors in the
training data~\cite{kearns93,frenay14}, only a few works have
considered adversarial, worst-case label noise, either from a more
theoretical~\cite{bshouty99} or practical
perspective~\cite{biggio11-acml,xiao12}.  In \cite{kearns93,bshouty99}
the impact of stochastic and adversarial label noise on the
classification error have been theoretically analyzed under the
\emph{probably-approximately-correct} learning model, deriving lower
bounds on the classification error as a function of the fraction of
flipped labels $\eta$; in particular, the test error can be shown to
be lower bounded by $\eta / (1-\eta)$ and $2\eta$ for stochastic and
adversarial label noise, respectively.  In recent
work~\cite{biggio11-acml,xiao12}, instead, we have focused on deriving
more practical attack strategies to maximize the test error of an SVM
given a maximum number of allowed label flips in the training data.
Since finding the worst label flips is generally computationally demanding, 
we have devised suitable heuristics to find
approximate solutions efficiently. To our knowledge, these are
the only works devoted to understanding how SVMs can be affected by
adversarial label noise.

From a more practical viewpoint, the problem is of interest as attackers may concretely have access and change some of the training labels in a number of cases. For instance, if feedback from end-users is exploited to label data and update the system, as in collaborative spam filtering, an attacker may have access to an authorized account (\eg, an email account protected by the same anti-spam filter), and manipulate the labels assigned to her samples. In other cases, a system may even ask directly to users to validate its decisions on some submitted samples, and use them to update the classifier (see, \eg, \verb=PDFRate=,\footnote{Available at: \url{http://pdfrate.com}} an online   tool for detecting PDF malware~\cite{smutz12}).
The practical relevance of poisoning attacks has also been recently discussed in the context of the detection of malicious crowdsourcing websites that connect paying users with workers willing to carry out malicious campaigns (\eg, spam campaigns in social networks) --- a recent phenomenon referred to as \emph{crowdturfing}. In fact, administrators of crowdturfing sites can intentionally pollute the training data used to learn classifiers, as it comes from
their websites, thus being able to launch poisoning attacks~\cite{wang14-usenix}.

In this paper, we extend our work on adversarial label noise against SVMs~\cite{biggio11-acml,xiao12} by improving our previously-defined attacks (Sects.~\ref{sect:alfa} and \ref{sect:alfa-tilt}), and by proposing two novel heuristic approaches. One has been inspired from previous work on SVM poisoning~\cite{biggio12-icml} and incremental learning~\cite{cauwenberghs00,diehl03}, and makes use of a continuous relaxation of the label values to greedily maximize the SVM's test error through gradient ascent (Sect.~\ref{sect:alfa-cr}).
The other exploits a breadth first search to greedily construct sets of candidate label flips that are \emph{correlated} in their effect on the test error (Sect.~\ref{sect:cc}).
As in~\cite{biggio11-acml,xiao12}, we aim at analyzing the maximum performance degradation incurred by an SVM under adversarial label noise, to assess whether these attacks can be considered a relevant threat. We thus assume that the attacker has perfect knowledge of the attacked system and of the training data, and left the investigation on how to develop such attacks having limited knowledge of the training data to future work. 
We further assume that the adversary incurs the same cost for flipping each label, independently from the corresponding data point.
We demonstrate the effectiveness of the proposed approaches by reporting experiments on synthetic and real-world datasets (Sect.~\ref{sect:exp}). We conclude in Sect.~\ref{sect:conclusions} with a discussion on the contributions of our work, its limitations, and future research, also related to the application of the proposed techniques to other fields, including semi-supervised and active learning.

\section{Support Vector Machines and Notation}
\label{sect:svm-notation}

We revisit here structural risk minimization and SVM learning,  and introduce the framework that will be used to motivate our attack
strategies for adversarial label noise.

In risk minimization, the goal is to find a hypothesis $\hyp: \set X \to
\set Y$ that represents an unknown relationship between an input  $\set X$ and an output space $\set Y$, captured by a probability measure $P$.
Given a non-negative \emph{loss function} $\lossChr : \set Y \times \set Y \to
\reals$ assessing the error between the prediction $\hat{y}$ provided by $\hyp$ and the true output $y$, we can define the optimal hypothesis $\hyp^\star$ as the one that minimizes the
expected risk $\risk{\hyp}{P} = \expect{(\vct
  x,y)\sim P}{\loss{\hyp(\vct x)}{y}}$ over the hypothesis space $\hypSpace$, \ie, $\hyp^\star = \argmin_{\hyp\in\hypSpace} \risk{\hyp}{P}$.
Although $P$ is not usually known, and thus $\hyp^\star$ can not be computed directly,  a set $\trainDS = \left\{ (\vct x_i, y_i) \right\}_{i=1}^{n}$ of i.i.d. samples drawn from $P$ are often available. In these cases a learning algorithm $\learnAlgo$ can be used to find a suitable hypothesis. 
According to structural risk minimization~\cite{vapnik95-book}, the learner $\learnAlgo$ minimizes a sum of a
regularizer and the empirical risk over the data:
\begin{align}
  \learnAlgo(\trainDS) = \argmin_{\hyp\in\hypSpace} \quad & \left[ \regularizer{\hyp} + \regParam \cdot \emprisk{\hyp}{\trainDS} \right] \, ,
  \label{eq:empRiskForm}
\end{align}
where the regularizer $\regularizer{\hyp}$ is used to penalize excessive  hypothesis complexity and avoid overfitting, the empirical
risk $\emprisk{\hyp}{\trainDS}$ is given by $\frac{1}{n} \sum_{i=1}^{n}
\loss{\hyp(\vct x_i)}{y_i}$, and $\regParam > 0$ is a parameter that controls the trade-off between minimizing the empirical loss and the complexity of
the hypothesis.

The SVM is an example of a binary linear classifier developed according to the aforementioned principle. It makes predictions in $\set Y = \{-1,+1\}$ based on the sign of its real-valued discriminant function $f(\vct x)=\vct w^{\T} \vct x +b$; \ie,  $\vct x$ is classified as positive if $f(\vct x) \geq 0$, and negative otherwise.
The SVM uses the hinge loss $\loss{\hyp(\vct x)}{y} = \max \left(0, 1 - y \hyp(\vct x) \right)$ as a convex surrogate loss function, and
a quadratic regularizer on $\vct w$, \ie,
$\regularizer{\hyp} = \frac{1}{2} \vct w^{\top} \vct w$.
Thus, SVM learning can be formulated according to
Eq.~\eqref{eq:empRiskForm} as the following convex quadratic
programming problem:
\begin{align}
  \label{eq:svm-primal}
  \min_{\vct w, b}\quad & \tfrac{1}{2}\vct w^{\T} \vct w + \regParam
    \tsum_{i=1}^{n} \max \left( 0,1-y_{i}f(\vct x_{i}) \right)    \enspace.
\end{align}

An interesting property of SVMs arises from their \emph{dual}
formulation, which only requires computing inner products between samples during training and classification, thus avoiding the need of an \emph{explicit} feature representation.
Accordingly, non-linear decision functions in the input space can be learned using \emph{kernels}, \ie, inner products in implicitly-mapped feature spaces. 
In this case, the SVM's decision
function is given as $f(\vct x)=\sum_{i=1}^{n} \alpha_{i} y_{i} k(\vct
x, \vct x_{i}) +b$, where $k(\vct x, \vct z)=\phi(\vct x)^{\T}
\phi(\vct z)$ is the kernel function, and $\phi$ the implicit mapping. The SVM's dual parameters $(\vctg \alpha,b)$ are found by solving the dual problem:
\begin{eqnarray}
\min_{0 \leq \vctg \alpha \leq C} & & \frac{1}{2}\vctg \alpha^{\T} \vct Q \vctg \alpha - \vct 1^{\T} \vctg \alpha 
\enspace , \enspace
{\rm s.t. } \enspace \vct y^{\T} \vctg \alpha = 0 \enspace ,
\end{eqnarray}
where $\vct Q = \vct y \vct y^{\T} \circ \vct K$ is the
label-annotated version of the (training) kernel matrix $\vct K$. The bias $b$ is obtained from the corresponding Karush-Kuhn-Tucker (KKT) conditions,
to satisfy the equality constraint $\vct y^{\T} \vctg \alpha = 0$ (see, \eg,~\cite{libsvm}). 

In this paper, however, we are not only interested in how the
hypothesis is chosen but also how it performs on a second validation
or test dataset $\valDS$, which may be generally drawn from a different
distribution $Q$. We thus define the error measure
\begin{align}
  \label{eq:flipping-obj}
  \regObj{\trainDS}{\valDS} & = \| \hyp_{\trainDS} \|^2 + \regParam \cdot
  \emprisk{\hyp_{\trainDS}}{\valDS} \enspace,
\end{align}
which implicitly uses $\hyp_{\trainDS} = \learnAlgo(\trainDS)$. This
function evaluates the structural risk of a
hypothesis $\hyp_{\trainDS}$ that is \emph{trained} on $\trainDS$ but
  \emph{evaluated} on $\valDS$, and will form the foundation for
our label flipping approaches to dataset poisoning.  Moreover, since
we are only concerned with label flips and their effect on the learner
we use the notation $\regObj{\vct z}{\vct y}$ to denote the above
error measure when the datasets differ only in the labels $\vct z$
used for training and $\vct y$ used for evaluation; \ie, $\regObj{\vct
  z}{\vct y} = \regObj{\;\{(\vct x_i, z_i)\}\;}{\;\{ (\vct x_i,
  y_i)\}\;}$.

\section{Adversarial Label Flips on SVMs}
\label{sect:alfa-svm}

In this paper we aim at gaining some insights on whether and to what
extent an SVM may be affected by the presence of
well-crafted mislabeled instances in the training data.  We assume the
presence of an attacker whose goal is to cause a denial of service, \ie, to maximize the SVM's
classification error, by changing the labels of at most $L$ samples
in the training set.  Similarly to \cite{xiao12,biggio12-icml}, the
problem can be formalized as follows.

We assume there is some learning algorithm
$\learnAlgo$ known to the adversary that maps a training dataset into
hypothesis space: $\learnAlgo : (\set X \times \set Y)^n \to
\hypSpace$. Although this could be any learning algorithm, we
consider SVM here, as discussed above.
The adversary wants to maximize the classification error (\ie, the risk), that the learner is trying to minimize, by contaminating the training
data so that the hypothesis is selected based on tainted data drawn
from an adversarially selected distribution $\hat{P}$ over $\set X
\times \set Y$. However, the adversary's capability of manipulating the training data is bounded by requiring $\hat{P}$ to be within a
neighborhood of (\ie, ``close
to'') the original distribution $P$.

For a worst-case label flip attack, the attacker is restricted to only
change the labels of training samples in $\trainDS$ and is allowed to
change at most $L$ such labels in order to maximally increase the
classification risk of $\learnAlgo$ --- $L$ bounds the attacker's
capability, and it is fixed \emph{a priori}. Thus, the problem can be
formulated as
\begin{align}
  \vct y^{\prime} = \argmax_{\vct z \in \set Y^n}\quad & \risk{\hyp_{\vct z}(\vct x)}{P} \enspace , \label{eq:attacker-obj}  \\
  \textrm{s.t.}\quad & \hyp_{\vct z} = \learnAlgo \left( \{ (\vct x_i, z_i) \}_{i=1}^{n} \right )  \enspace , \nonumber \\
& \tsum_{i=1}^{n} \ind{z_i \neq y_i} \le L \nonumber
\enspace ,
\end{align}
where $\ind$ is the indicator function, which returns one if the argument is true, and zero otherwise.
However, as with the learner, the true risk $\riskChr$ cannot be
assessed because $P$ is also unknown to the adversary. As with the
learning paradigm described above, the risk used to select $\vct
y^{\prime}$ can be approximated using by the regularized empirical
risk with a convex loss.  Thus the objective in
Eq.~\eqref{eq:attacker-obj} becomes simply $\regObj{\vct z}{\vct y}$
where, notably, the empirical risk is measured with respect to the
true dataset $\trainDS$ with the original labels. For the SVM and
hinge loss, this yields the following program:
\begin{align}
  \label{eq:svm-attack-obj}
  \max_{\vct z \in\{-1,+1\}^{n}}\quad & \tfrac{1}{2}\vctg \alpha^{\T} \vct Q \vctg \alpha + \regParam \tsum_{i=1}^{n} \max(0,1-y_{i} \hyp_{\vct z}(\vct x_i)) \enspace , \\
   \textrm{s.t.}\quad & (\vctg \alpha,b) = \learnAlgo_{SVM} \left (\{ (\vct x_i, z_i) \}_{i=1}^{n} \right ) \enspace , \nonumber \\
& \tsum_{i=1}^{n} \ind{z_i \neq y_i} \le L \nonumber
\enspace ,
\end{align}
where $\hyp_{\vct z}(\vct x) = \tsum_{j=1}^{n} z_j \alpha_j K(\vct
x_{j},\vct x) + b$ is the SVM's dual discriminant function learned
on the tainted data with labels $\vct z$.

The above optimization is a $\mathcal{NP}$-hard
subset selection problem, that includes SVM learning
as a subproblem. In the next sections we present a set
of heuristic methods to find approximate solutions to the posed problem
efficiently;
in particular, in Sect.~\ref{sect:alfa} we revise the approach proposed in \cite{xiao12} according to the aforementioned framework, in Sect.~\ref{sect:alfa-cr} we present a novel approach for adversarial label flips based on a continuous relaxation of Problem~\eqref{eq:svm-attack-obj}, in Sect.~\ref{sect:alfa-tilt} we present an improved, modified version of the approach we originally proposed in \cite{biggio11-acml}, and in Sect.~\ref{sect:cc} we finally present another, novel approach for adversarial label flips that aims to flips clusters of labels that are `correlated' in their effect on the objective function.

\subsection{Adversarial Label Flip Attack (\textit{alfa})}
\label{sect:alfa}

We revise here the near-optimal label flip attack proposed in \cite{xiao12}, named Adversarial Label Flip Attack (\texttt{alfa}).
It is formulated under the assumption that the attacker can maliciously manipulate the set of labels to maximize the empirical loss of the original classifier on
the tainted dataset, while the classification algorithm preserves its
generalization on the tainted dataset without noticing it.  The
consequence of this attack misleads the classifier to an erroneous shift
of the decision boundary, which most deviates from the untainted original data
distribution.

As discussed above, given the untainted dataset $\trainDS$ with $n$ labels
$\vct y$, the adversary aims to flip at most $L$ labels to form the
tainted labels $\vct z$ that maximize $\regObj{\vct z}{\vct y}$.
Alternatively, we can pose this problem as a search for labels $\vct z$ that achieve the maximum difference between the empirical risk for classifiers trained on $\vct z$ and $\vct y$, respectively. The attacker's objective can thus be expressed as
\begin{align}
  \min_{\vct z \in\{-1,+1\}^{n}}\quad & \regObj{\vct z}{\vct z}-\regObj{\vct y}{\vct z} \enspace ,\label{eq:obj-ad2}\\
  \mathrm{s.t.} \quad & \tsum_{i=1}^{n} \ind{z_i \neq y_i} \le L \nonumber
  \enspace.
\end{align}

To solve this problem, we note that the $\emprisk{\hyp}{\valDS}$
component of $\objChr$ is a sum of losses over the data points, and the
evaluation set $\valDS$ only differs in its labels. Thus, for each
data point, either we will have a component $\loss{\hyp(\vct x_i)}{y_i}$ or
a component $\loss{\hyp( \vct x_i)}{-y_i}$ contributing to the risk. 
By denoting with an indicator variable $q_i$ which component to use,
the attacker's objective can be rewritten as the problem of minimizing the following expression with respect to $\vct q$ and $\hyp$:
\[
\| \hyp \|^{2}+ \regParam \sum_{i=1}^{n} \begin{matrix} 
    (1-q_{i}) \cdot [ \loss{\hyp(\vct{x}_{i})}{y_{i}} - \loss{\hyp_{\vct y}(\vct{x}_{i})}{y_{i}}] \\
    + q_{i} \cdot [ \loss{\hyp(\vct{x}_{i})}{-y_{i}} - \loss{\hyp_{\vct y}(\vct{x}_{i})}{-y_{i}}]
    \end{matrix}
  \enspace.
\]
In this expression, the dataset is
effectively duplicated and either $(\vct x_i,y_i)$ or $(\vct x_i,
-y_i)$ are selected for the set $\valDS$.  The $q_i$ variables are
used to select an optimal subset of labels $y_i$ to be flipped for
optimizing $\hyp$.

When \texttt{alfa} is applied to the SVM, we use the hinge loss and the primal
SVM formulation from Eq.~\eqref{eq:svm-primal}. 
We denote with $\xi_i^{0} = \max(0,1-y_{i}\hyp_{\vct y}(\vct x_{i}))$ 
and $\xi_i^{1} = \max(0,1+y_{i}\hyp_{\vct y}(\vct x_{i}))$ the
\nth{i}{th} loss of classifier $\hyp_{\vct y}$ when the
\nth{i}{th} label is respectively kept unchanged or flipped. Similarly,
$\epsilon_i^{0}$ and $\epsilon_i^{1}$ are the corresponding slack
variables for the new classifier $\hyp$.
The above attack framework can then be expressed as:
\begin{align}
  \min_{\vct q,\vct w, \vctg \epsilon^0, \vctg \epsilon^1, b}\quad&
  \frac{1}{2}\|\vct w\|^{2} + \regParam\sum_{i=1}^{n} \begin{bmatrix}
    (1-q_{i})\cdot(\epsilon_{i}^{0}-\xi_{i}^{0}) \\
    + q_{i}\cdot(\epsilon_{i}^{1}-\xi_{i}^{1})
  \end{bmatrix} \enspace , \label{eq:svm-attack}
  \\
  \mathrm{s.t.} \quad 
  & 1-\epsilon_{i}^{0} \leq y_{i}(\vct w^{\top}\vct x_{i}+b) \leq \epsilon_{i}^{1}-1,\quad \epsilon_{i}^{0},\epsilon_{i}^{1} \geq 0 \enspace , \nonumber \\
   & \tsum_{i=1}^{n}q_{i} \leq L, \quad q_i \in \{0,1\} \nonumber
\enspace.
\end{align}
To avoid integer programming which is generally $\mathcal{NP}$-hard,
the indicator variables, $q_i$, are relaxed to be continuous on
$\left[0, 1\right]$.
The minimization problem in Eq.~\eqref{eq:svm-attack} is then decomposed
into two iterative sub-problems. First, by fixing $\vct q$, the
summands $\xi_i^{0} + q_i (\xi_i^{0} - \xi_i^{1})$ are constant, and
thus the minimization reduces to the following QP problem:
\begin{align}
  \min_{\vct w, \vctg \epsilon^{0}, \vctg \epsilon^{1}, b}\quad&
  \frac{1}{2}\|\vct w\|^{2}+\regParam\sum_{i=1}^{n} \left[ (1-q_{i})\epsilon_{i}^{0} + q_i\epsilon_i^{1}\right]  \enspace ,
  \label{eq:part1-qp}
  \\
  \mathrm{s.t.} \quad 
  & 1-\epsilon_{i}^{0} \leq y_{i}(\vct w^{\top}\vct x_{i}+b) \leq \epsilon_{i}^{1}-1,\quad \epsilon_{i}^{0},\epsilon_{i}^{1} \geq 0 \enspace . \nonumber
\end{align}
Second, fixing $\vct w$ and $b$ yields a set of fixed hinge losses,
$\vctg \epsilon^0$ and $\vctg \epsilon^1$.  The minimization over
(continuous) $\vct q$ is then a linear programming problem (LP):
\begin{align}
  \min_{\vct q \in [0,1]^n}\quad&
  \regParam\sum_{i=1}^{n} (1-q_{i})(\epsilon_{i}^0-\xi_{i}^0) + q_{i}(\epsilon_{i}^1-\xi_{i}^1) \enspace ,
  \label{eq:part2-lp}
  \\
  \textrm{s.t.} \quad &\tsum_{i=1}^{n}q_{i} \leq L \nonumber
  \enspace.
\end{align}
After convergence of this iterative approach, the largest subset of
$\vct q$ corresponds to the near-optimal label flips within the budget $L$.  The complete \texttt{alfa} procedure is given as Algorithm~\ref{alg:alfa}.

\begin{algorithm}[tb]
  \SetKwInOut{Input}{Input}\SetKwInOut{Output}{Output}
\SetKwFunction{Sort}{Sort}
\SetKwData{Ind}{Ind}
\Input{Untainted training set $\trainDS=\{\vct x_{i}, y_{i}\}_{i=1}^{n}$, maximum number of label flips $L$.}
\Output{Tainted training set $\trainDS^{\prime}$.}

Find $\hyp_{\vct y}$ by solving \cref{eq:svm-primal} on $\trainDS$ \tcc*[r]{QP}
\ForEach{$(\vct x_{i},y_{i})\in \trainDS$}{
  $\xi_{i}^0\leftarrow\max(0,1-y_{i}\hyp_{\vct y}(\vct x_{i}))$\;
  $\xi_{i}^1\leftarrow\max(0,1+y_{i}\hyp_{\vct y}(\vct x_{i}))$\;
  $\epsilon_{i}^0\leftarrow 0$, and $\epsilon_{i}^1\leftarrow 0$\;
}
\Repeat{convergence}{
Find $\vct q$ by solving \cref{eq:part2-lp}\tcc*[r]{LP}
Find $\vctg \epsilon^{0},\vctg \epsilon^{1}$ by solving \cref{eq:part1-qp}\tcc*[r]{QP}
}
$\mathit{v}\leftarrow$\Sort{$[q_{1},\ldots,q_{n}]$, ``{\rm descend}''}\;
\tcc{$\mathit{v}$ are sorted indices from $n+1$}
\lFor{$i\leftarrow 1$ \KwTo $n$}{$z_{i}\leftarrow y_{i}$}
\lFor{$i\leftarrow 1$ \KwTo $L$}{$z_{\mathit{v}[i]}\leftarrow
  -y_{\mathit{v}[i]}$}
\Return $\trainDS^{\prime}\leftarrow\{(\mathbf{x}_{i},z_{i})\}_{i=1}^{n}$\;
  \caption{\texttt{alfa}}
  \label{alg:alfa}
\end{algorithm}

\subsection{ALFA with Continuous Label Relaxation (\textit{alfa-cr})}
\label{sect:alfa-cr}

The underlying idea of the method presented in this section is to
solve Problem~\eqref{eq:svm-attack-obj} using a continuous relaxation
of the problem. In particular, we relax the constraint that the
tainted labels $\vct z \in \{-1,+1\}^{n}$ have to be
discrete, and let them take on continuous real values on a bounded domain.
We thus maximize the objective function in Problem~\eqref{eq:svm-attack-obj} with respect to $\vct
z \in [z_{\rm min}, z_{\rm max}]^{n} \subseteq \mathbb R^{n}$.
Within this assumption, we optimize the
objective through a simple gradient-ascent algorithm, and
iteratively map the continuous labels to discrete values during the
gradient ascent. The gradient derivation and the complete attack
algorithm are respectively reported in
Sects.~\ref{sect:alfa-cr-gradient} and~\ref{sect:alfa-cr-algorithm}.

\subsubsection{Gradient Computation} 
\label{sect:alfa-cr-gradient}

Let us first compute the gradient of the objective in Eq.~\eqref{eq:svm-attack-obj}, starting from the loss-dependent term $\sum_{i} \max \left (0,1-y_{i}f_{\vct z}(\vct x_{i}) \right )$. Although this term is not differentiable when $y f_{\vct z}(\vct x)=1$, it is  possible to consider a subgradient that is equal to the gradient of  $1-y f_{\vct z}(\vct x)$, when $y f_{\vct z}(\vct x)<1$, and to $0$ otherwise.
The gradient of the loss-dependent term is thus given as:
\begin{equation}
\label{eq:grad_V}
\diff{}{\vct z}\left( \sum_{i} \max \left (0,1-y_{i}f_{\vct z}(\vct x_{i}) \right ) \right )
= -\sum_{i =1}^{n}   \delta_{i}  \diff{ v_{i}}{\vct z} \enspace ,
\end{equation}
where $\delta_{i}=1$ (0) if $y_{i}f_{\vct z}(\vct x_{i}) < 1$ (otherwise), and
\begin{align}
   v_{i} = y_{i} \left(  \sum_{j=1}^{n} K_{ij} z_{j}(\vct z) \alpha_j (\vct z) + b(\vct z) \right ) -1 \; ,
    \label{eq:fk_def}
\end{align}
where we explicitly account for the dependency on $\vct z$. 
To compute the gradient of $v_{i}$, we derive this expression with respect to each label $z_{\ell}$ in the training data using the product rule:
\begin{eqnarray}
\diff{v_{i}}{z_{\ell}} = y_{i} \left ( \sum_{j=1}^{n} K_{ij} z_{j} \diff{\alpha_{j}}{z_{\ell}} + K_{i \ell} \alpha_{\ell} + \diff{b}{z_{\ell}} \right ) \enspace . 
\end{eqnarray}
%
This can be compactly rewritten in matrix form as:
\begin{eqnarray}
\label{eq:gradient-gk}
\diff{\vct v}{\vct z} = \left (\vct y\vct z^{\T}  \circ \vct K \right ) \diff{\vctg \alpha}{\vct z} + \vct K \circ (\vct y \vctg \alpha^{\T}) + \vct y \diff{b}{\vct z} \enspace ,
\end{eqnarray}
where, using the numerator layout convention, 
\begin{equation*}
  \diff{\vctg \alpha}{\vct z} = 
  \begin{bmatrix}
    \diff{\alpha_1}{z_1} & \cdots & \diff{\alpha_1}{z_{n}} \\
    \vdots & \ddots & \vdots\\
    \diff{\alpha_n}{z_1} & \cdots & \diff{\alpha_n}{z_n}
  \end{bmatrix}, \; 
  \diff{b}{\vct z}^{\T} = \begin{bmatrix} \diff{b}{z_{1}} \\ \cdots \\ \diff{b}{z_{n}} \end{bmatrix} , \; {\rm and \; simil.} \; \diff{\vct v}{\vct z} \enspace .
\end{equation*}

The expressions for $\diff{\vctg \alpha}{\vct z}$ and $\diff{b}{\vct z}$ required to compute the gradient in Eq.~\eqref{eq:gradient-gk} can be obtained by assuming that the SVM solution remains in equilibrium while $\vct z$ changes smoothly. This can be expressed as an adiabatic update condition using the technique introduced in \cite{cauwenberghs00,diehl03}, and exploited in \cite{biggio12-icml} for a similar gradient computation. Observe that for the \emph{training} samples, the KKT conditions for the optimal solution of the SVM training problem can be expressed as:
\begin{align}
\vct g &=   \vct Q  \vctg \alpha+ \vct z b-1
\begin{cases}
{\rm if} \; g_{i}> 0, \;  i \in \mathcal R\\
{\rm if} \; g_{i}= 0, \;  i \in \mathcal S \\
{\rm if} \; g_{i}< 0, \;  i \in \mathcal E
\end{cases} \label{eq:kt-svm-1}\\
h &=  \vct z^{ \T} \vctg \alpha=0 \label{eq:kt-svm-2} \enspace ,
\end{align}
where we remind the reader that, in this case, $\vct Q = \vct z\vct z^{\T}  \circ \vct K$.
The equality in condition \eqref{eq:kt-svm-1}-\eqref{eq:kt-svm-2} implies that an infinitesimal change in $\vct z$ causes a smooth change in the optimal solution of the SVM, under the constraint that the sets $\mathcal{R}$, $\mathcal{S}$, and $\mathcal{E}$ do not change. This allows us to predict the \emph{response} of the SVM solution
to the variation of $\vct z$ as follows.

By differentiation of Eqs.~\eqref{eq:kt-svm-1}-\eqref{eq:kt-svm-2}, we obtain:
\begin{eqnarray}
\diff{\vct g}{\vct z}& =& \vct Q \diff{\vctg \alpha}{\vct z} + \vct K \circ (\vct z \vctg \alpha^{\T}) + \vct z \diff{b}{\vct z}  + \vct S \, , \\
\diff{h}{\vct z}^{\T} &=& \vct z^{\T} \diff{\vctg \alpha}{\vct z} +\vctg \alpha^{\T} \, ,
\end{eqnarray}
where $\vct S={\rm diag}(\vct K (\vct z \circ \vctg \alpha) + b)$ is an $n$-by-$n$ diagonal matrix, whose elements $S_{ij}=0$ if $i \neq j$, and $S_{ii}=f_{\vct z}(\vct x_{i})$ elsewhere. 

The assumption that the SVM solution does not change structure while updating $\vct z$ implies that
\begin{eqnarray}
\diff{\vct g_{s}}{\vct z}= 0 , \enspace  \diff{h}{\vct z} =0 ,
\end{eqnarray}
where $s$ indexes the \emph{margin} support vectors in $\set S$ (from the equality in condition \ref{eq:kt-svm-1}).
In the sequel, we will also use $r$, $e$, and $n$, respectively to index the \emph{reserve} vectors in $\set R$, the \emph{error} vectors in $\set E$, and all the $n$ training samples.
The above assumption leads to the following linear problem, which allows us to predict how the SVM solution changes while $\vct z$ varies:
\begin{equation}
\label{eq:lin_prob}
\begin{bmatrix}
\vct Q_{ss} & \vct z_{s} \\
\vct z^{\T}_{s} & 0 \\
\end{bmatrix}
\begin{bmatrix}
\diff{\vctg \alpha_{s}}{\vct z} \\
\diff{b}{\vct z}
\end{bmatrix}
=
-
\begin{bmatrix}
\vct K_{sn} \circ (\vct z_{s} \vctg \alpha^{\T})  + \vct S_{sn} \\
\vctg \alpha^{\T}
\end{bmatrix} \enspace .
\end{equation}
The first matrix can be inverted using matrix block inversion~\citep{lue96}:
\begin{equation}
  \label{eq:smw}
  \begin{aligned}
   	\begin{bmatrix}
		\vct Q_{ss} & \vct z_{s} \\
		\vct z^{\T}_{s} & 0 \\
	\end{bmatrix}^{-1}
    = \frac{1}{\zeta}
    \begin{bmatrix}
      \zeta \vct Q_{ss}^{-1} -\vctg \upsilon \vctg \upsilon^{\T} & \vctg \upsilon \\
      \vctg \upsilon^{\T} & -1
    \end{bmatrix} \, ,
  \end{aligned}
\end{equation}
where $\vctg \upsilon = \vct Q_{ss}^{-1} \vct z_s$ and $\zeta = \vct z^{\T}_s \vct Q_{ss}^{-1}
\vct z_s$. Substituting this result to solve Problem~\eqref{eq:lin_prob}, one obtains:
\begin{align}
\label{eq:alpha_s}
\diff{\vctg \alpha_{s}}{\vct z}  &=  \left (\frac{1}{\zeta}\vctg \upsilon \vctg \upsilon^{\T}- \vct Q_{ss}^{-1}\right)\left (\vct K_{sn} \circ (\vct z_{s} \vctg \alpha^{\T})  + \vct S_{sn}\right )  - \frac{1}{\zeta}\vctg \upsilon \vctg \alpha^{\T} \, ,\\
\label{eq:b}
\diff{b}{\vct z} &= - \frac{1}{\zeta} \vctg \upsilon^{\T} \left( \vct K_{sn} \circ (\vct z_{s} \vctg \alpha^{\T})  + \vct S_{sn} \right) + \frac{1}{\zeta} \vctg \alpha^{\T} \, .
\end{align}

The assumption that the structure of the three sets $\set S, \set R, \set E$ is preserved also implies that $\diff{\vctg \alpha_{r}}{\vct z}= \vct 0$ and $\diff{\vctg \alpha_{e}}{\vct z}=\vct 0$. Therefore, the first term in Eq.~\eqref{eq:gradient-gk} can be simplified as:
\begin{align}
\label{eq:gradient-gk-simplified}
\diff{\vct v}{\vct z} = \left (\vct y\vct z_{s}^{\T}  \circ \vct K_{ns} \right ) \diff{\vctg \alpha_{s}}{\vct z} + \vct K \circ (\vct y \vctg \alpha^{\T}) + \vct y \diff{b}{\vct z} \enspace .
\end{align}

Eqs.~\eqref{eq:alpha_s} and \eqref{eq:b} can be now substituted into Eq.~\eqref{eq:gradient-gk-simplified}, and further into Eq.~\eqref{eq:grad_V} to compute the gradient of the loss-dependent term of our objective function.

As for the regularization term, the gradient can be simply computed as: 
\begin{equation}
\diff{}{\vct z} \left(\frac{1}{2}\vctg \alpha^{\T} \vct Q \vctg \alpha \right )= \vctg \alpha \circ [\vct K (\vctg \alpha \circ \vct z) ] + \diff{\vctg \alpha}{\vct z}^{\T} \vct Q \vctg \alpha \enspace .
\end{equation}
Thus, the complete gradient of the objective in Problem~\eqref{eq:svm-attack-obj} is:
\begin{equation}
\nabla_{\vct z} \objChr=\vctg \alpha \circ [\vct K (\vctg \alpha \circ \vct z) ] + \diff{\vctg \alpha}{\vct z}^{\T} \vct Q \vctg \alpha - C  \sum_{i =1}^{n}   \delta_{i}  \diff{ v_{i}}{\vct z} \enspace.
\end{equation}

The structure of the SVM (\ie, the sets $\set S, \set R, \set E$) will clearly change while updating $\vct z$, hence after each gradient step we should re-compute the optimal SVM solution along with its corresponding structure.
This can be done by re-training the SVM from scratch at each iteration.
Alternatively, since our changes are \emph{smooth}, the SVM solution can be more efficiently updated at each iteration using an active-set optimization algorithm initialized with the $\vctg \alpha$ values obtained from the previous iteration as a warm start~\cite{scheinberg06}. Efficiency may be further improved by developing an ad hoc incremental SVM under label perturbations based on the above equations. This however includes the development of suitable bookkeeping conditions, similarly to \cite{cauwenberghs00,diehl03}, and it is thus left to future investigation.

\subsubsection{Algorithm}
\label{sect:alfa-cr-algorithm}

Our attack algorithm for \verb=alfa-cr= is given as Algorithm~\ref{alg:alfa-cr}. It exploits the gradient derivation reported in the previous section to maximize the objective function $\objChr(\vct z,\vct y)$ with respect to continuous values of $\vct z \in [z_{\rm min}, z_{\rm max}]^{n}$.  The current best set of continuous labels is iteratively mapped to the discrete set $\{-1,+1\}^{n}$, adding a label flip at a time, until $L$ flips are obtained.

\begin{algorithm}[t]
  \SetKwInOut{Input}{Input}\SetKwInOut{Output}{Output}
\SetKwFunction{Sort}{Sort}
\SetKwData{Ind}{Ind}
\Input{Untainted training set $\trainDS=\{\vct x_{i}, y_{i}\}_{i=1}^{n}$, maximum num. of label flips $L$, maximum num. of iterations $N$ ($N\geq L$), gradient step size $t$.}
\Output{Tainted training set $\trainDS^{\prime}$.}

$\vct z \leftarrow \vct y$ \tcc*[r]{Initialize labels}
$\vct z_{\rm best} \leftarrow \vct y$ \;
$\{\vctg \alpha, b\} \leftarrow$ learn SVM on $\trainDS$ (Eq.~\ref{eq:svm-primal}) \;
$p \leftarrow 0$ \tcc*[r]{Number of current flips}
$k \leftarrow 0$ \tcc*[r]{Number of iterations}
\While{$p < L$}{
	$k \leftarrow k+1$ \;
	\tcc{Compute gradient from \cref{eq:grad_V} using current SVM solution}
	$\vct z \leftarrow \vct z + t \nabla_{\vct z}{\objChr}$ \;
	Project $\vct z$ onto the feasible domain $[z_{\rm min}, z_{\rm max}]^{n}$ \;
	$\{\vctg \alpha, b\} \leftarrow$ update SVM solution on $\{\vct x_{i}, z_{i}\}_{i=1}^{n}$ \;
	\If{$\objChr(\vct z,\vct y) \geq \objChr(\vct z_{\rm best},\vct y) $}{
		$\vct z_{\rm best} \leftarrow \vct z$ \tcc*[r]{Best (soft) labels}
	}
	\If{$\mod(k,  \lfloor N/L \rfloor )=0$}{
		\tcc{Project the best (soft) labels to $p$ (hard) label flips}
		$p \leftarrow p +1$ \tcc*[r]{Update flip count}
		$\vct z \leftarrow$ Flip the first $p$ labels from $\vct y$ according to the descending order of $|\vct z_{\rm best}-\vct y|$\;
	}
}
\Return $\set D^{\prime}_{\rm tr} \leftarrow\{(\mathbf{x}_{i},z_{i})\}_{i=1}^{n}$\;
  \caption{\texttt{alfa-cr}}
  \label{alg:alfa-cr}
\end{algorithm}

\subsection{ALFA based on Hyperplane Tilting (\textit{alfa-tilt})}
\label{sect:alfa-tilt}

\begin{algorithm}[t]
  \SetKwInOut{Input}{Input}\SetKwInOut{Output}{Output}
\SetKwFunction{Sort}{Sort}
\SetKwData{Ind}{Ind}
\Input{Untainted training set $\trainDS = \{ \mathbf x_{i}, y_{i} \}_{i=1}^{n}$, maximum num. of label flips $L$, maximum num. of trials $N$, and weighting parameters $\beta_{1}$ and $\beta_{2}$.}
\Output{Tainted training set $\trainDS^{\prime}$.}

$\{\vctg \alpha, b\} \leftarrow$ learn SVM on $\trainDS$ (Eq.~\ref{eq:svm-primal}) \;
\For{$i=1,\ldots,n$}{
	$s_{i} \gets y_{i}[\sum_{j=1}^{n} y_{j} \alpha_{j} K(x_{i},x_{j}) + b]$
	}
normalize $\{s_{i}\}_{i=1}^{n}$ dividing by $\max_{i=1,\ldots,n} s_{i}$ \;
$(\vctg \alpha^{\rm rnd}$, $b^{\rm rnd}) \gets$ generate a random SVM (draw $n+1$ numbers from a uniform distribution) \;
\For{ $i=1,\ldots,n$}{
	$q_{i} \gets y_{i}[\sum_{j=1}^{n} y_{j} \alpha^{\rm rnd}_{j} K(x_{i},x_{j}) + b^{\rm rnd}]$
	}
normalize $\{q_{i}\}_{i=1}^{n}$ dividing by $\max_{i=1,\ldots,n} q_{i}$ \;
\For{$i = 1,\ldots,n$}{$v_{i} \gets \alpha_{i}/C -  \beta_{1} s_{i} - \beta_{2} q_{i}$}
$(k_{1}, \ldots, k_{n}) \gets$ sort $(v_{1}, \ldots, v_{n})$ in ascending order, and return the corresponding indexes \;
$\vct z \gets \vct y$ \;
\For{ $i = 1,\ldots,L$}{ $z_{k_{i}} = -z_{k_{i}}$}
train an SVM on $\{ \mathbf x_{i}, z_{i} \}_{i=1}^{n}$ \;
estimate the hyperplane tilt angle using Eq.~\eqref{eq:alfa-tilt-obj} \;
repeat $N$ times from 5, selecting $\vct z$ to maximize Eq.~\eqref{eq:alfa-tilt-obj} \;
\Return{$\trainDS^{\prime} \gets \{\vct x_{i}, z_{i}\}_{i=1}^{n}$} \;
\caption{\texttt{alfa-tilt}~\cite{biggio11-acml}}
\label{alg:alfa-tilt}
\end{algorithm}

We now propose a modified version of the adversarial label flip attack we presented in~\cite{biggio11-acml}. The underlying idea of the original strategy is to generate different candidate sets of label flips according to a given heuristic method (explained below), and retain the one that maximizes the test error, similarly to the objective of Problem~\eqref{eq:svm-attack-obj}. 
However, instead of maximizing the test error directly, here we consider a surrogate measure, inspired by our work in~\cite{nelson11-aisec}. In that work, we have shown that, under the \emph{agnostic} assumption that the data is uniformly distributed in feature space, the SVM's robustness against label flips can be related to the change in the angle between the hyperplane $\vct w$ obtained in the absence of attack, and that learnt on the tainted data with label flips $\vct w^{\prime}$. Accordingly, the \verb=alfa-tilt= strategy considered here, aims to maximize the following quantity:
\begin{equation}
\max_{\vct z \in \{ -1,+1\}^{n}} \frac{\langle \vct w^{\prime}, \vct w \rangle}{||\vct w^{\prime}||||\vct w||} = \frac{\vctg \alpha^{\prime \T} \vct Q_{\vct z \vct y}  \vctg \alpha}{\sqrt{ \vctg \alpha^{\prime \T} \vct Q_{\vct z \vct z} \vctg \alpha^{\prime}} \sqrt{ \vctg \alpha^{\T} \vct Q_{\vct y \vct y} \vctg \alpha}} \enspace ,
\label{eq:alfa-tilt-obj}
\end{equation}
where $\vct Q_{\vct u \vct v} = \vct K \circ (\vct u \vct v^{\T})$, being $\vct u$ and $\vct v$ any two sets of training labels, and $\vctg \alpha$ and $\vctg \alpha^{\prime}$ are the SVM's dual coefficients learnt from the untainted and the tainted data, respectively.  

Candidate label flips are generated as explained in~\cite{biggio11-acml}.
Labels are flipped with non-uniform probabilities, depending on how well the corresponding training samples are classified by the SVM learned on the untainted training set. We thus increase the probability of flipping labels of reserve vectors (as they are reliably classified), and decrease the probability of label flips for margin and error vectors (inversely proportional to $\alpha$). The former are indeed more likely to become margin or error vectors in the SVM learnt on the tainted training set, and, therefore, the resulting hyperplane will be closer to them.
This will in turn induce a significant change in the SVM solution, and, potentially, in its test error.
We further flip labels of samples in different classes in a correlated way to force the hyperplane to rotate as much as possible. To this aim, we draw a random hyperplane $\vct w_{\rm rnd}$, $b_{ \rm rnd}$ in feature space, and further increase the probability of flipping the label of a positive sample $\vct x^{+}$ (respectively, a negative one $\vct x^{-}$), if $\vct{w}_{\rm rnd}^{\T} \vct{x}^{+} + b_{\rm rnd} > 0$ ($\vct{w}_{\rm rnd}^{\T} \vct{x}^{-} + b_{\rm rnd} < 0$).

The full implementation of \verb=alfa-tilt= is given as Algorithm~\ref{alg:alfa-tilt}. It depends on the parameters $\beta_{1}$ and $\beta_{2}$, which tune the probability of flipping a point's label based on how well it is classified, and how well it is correlated with the other considered flips.
As suggested in \cite{biggio11-acml}, they can be set to $0.1$, since this configuration has given reasonable results on several datasets.

\subsection{Correlated Clusters}
\label{sect:cc}

Here, we explore a different approach to heuristically optimizing
$\regObj{\vct z}{\vct y}$ that uses a breadth first search to greedily
construct subsets (or \emph{clusters}) of label flips that are `correlated'
in their effect on $\objChr$. Here, we use the term \emph{correlation}
loosely. 

\begin{algorithm}[t]
  \SetKwInOut{Input}{Input}\SetKwInOut{Output}{Output}
  \SetKwFunction{Sort}{Sort}
  \SetKwData{Ind}{Ind}
  \Input{Untainted training set $\trainDS=\{\vct x_{i}, y_{i}\}_{i=1}^{n}$, maximum number of label flips $L$, maximum number of iterations $N$ ($N\geq L$).}
  \Output{Tainted training set $\trainDS^{\prime}$.}

  Let $err(\vct z) = \emprisk{\learnAlgo(\{ (\vct x_i, z_i \})}{\trainDS}$\;
  $\hyp_{\vct y} \gets \learnAlgo(\trainDS)$, 
  $E_{\vct y} \gets err(\vct y)$\;
  $E^{\star} \gets -\infty$, $\vct z^{\star} \gets \vct y$\;
  \tcc{Choose random singleton clusters}
\For{i=1\ldots M}{
    $j \gets rand(1,n)$\;
    $\vct z^i \gets flip(\vct y,j)$\;
    $E_{i} \gets err(\vct z^i) - E_{\vct y}$\;
    \lIf{$E_{i} > E^{\star}$}{$E^{\star} \gets E_i$, $\vct z^{\star} \gets \vct z^i$}
    \For{j=1\ldots n} {
      \lIf{$rand_{[0,1]} < L/n$}{$\Delta_{i,j} \gets err(flip(\vct z^i,j))$}
      \lElse{$\Delta_{i,j} \gets -\infty$}
    }
  }
  \tcc{Grow new clusters by mutation}
 \For{t=1 \ldots N}{ 
    $(i,j) \gets \argmax_{(i,j)} \Delta_{i,j}$
    $\Delta_{i,j} \gets -\infty$
    $\vct z^{M+1} \gets flip(\vct z^i,j)$\;
    \If{$\|\vct z^{M+1} - \vct y \|_1 > 2L$}{Find best flip to reverse and flip it}
    $E_{M+1} \gets err(\vct z^{M+1}) - E_{\vct y}$\;
    \lIf{$E_{M+1} > E^{\star}$}{$E^{\star} \gets E_{M+1}$, $\vct z^{\star} \gets \vct z^{M+1}$}
    \For{k=1\ldots n} {
      \lIf{$rand_{[0,1]} < L/n$}{$\Delta_{M+1,k} \gets err(flip(\vct z^{M+1},k))$}
      \lElse{$\Delta_{M+1,k} \gets -\infty$}
    }
    Delete worst cluster and its entries in $E$ and $\Delta$\;
  }
\Return $\set D^{\prime}_{\rm tr} \leftarrow\{(\mathbf{x}_{i},z_{i}^{\star})\}_{i=1}^{n}$\;
  \caption{\texttt{correlated-clusters}}
  \label{alg:corr-clusters}
\end{algorithm}

The algorithm starts by assessing how each singleton flip impacts
$\objChr$ and proceeds by randomly sampling a set of $P$ initial
singleton flips to serve as initial clusters. For each of these
clusters, $k$, we select a random set of mutations to it (\ie, a
mutation is a change to a single flip in the cluster), which we then
evaluate (using the empirical 0-1 loss) to form a matrix
$\Delta$. This matrix is then used to select the best mutation to make
among the set of evaluated mutations.  Clusters are thus grown to
maximally increase the empirical risk.

To make the algorithm tractable, the population of candidate clusters
is kept small.  Periodically, the set of clusters are pruned to keep
the population to size $M$ by discarding the worst evaluated clusters.
Whenever a new cluster achieves the highest empirical error, that
cluster is recorded as being the best candidate cluster.  Further, if
clusters grow beyond the limit of $L$, the best \emph{deleterious}
mutation is applied until the cluster only has $L$ flips. This overall
process of greedily creating clusters with respect to the best
observed random mutations continues for a set number of iterations $N$
at which point the best flips until that point are returned.
Pseudocode for the correlated clusters algorithm is given in
Algorithm~\ref{alg:corr-clusters}.

\section{Experiments}
\label{sect:exp}

We evaluate the adversarial effects of various attack strategies against SVMs on both synthetic and real-world datasets. 
Experiments on synthetic datasets provide a conceptual representation of the rationale according to which the proposed attack strategies select the label flips. 
Their effectiveness, and the security of SVMs against adversarial label flips, is then more systematically assessed on different real-world datasets.
 
\subsection{On Synthetic Datasets}

\begin{figure*}[t]
\centering
\includegraphics[width=0.99\textwidth]{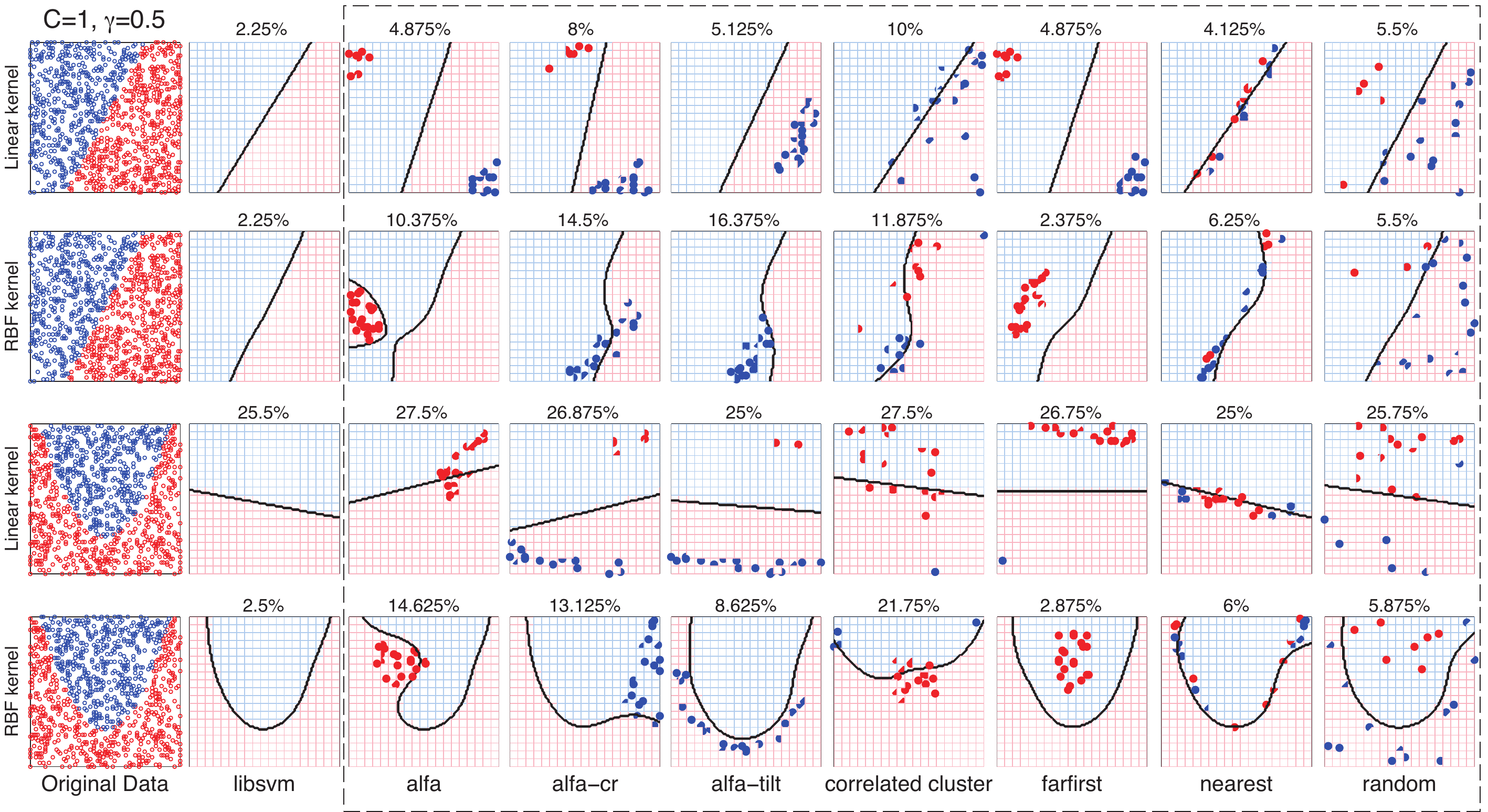}
\caption{Results on synthetic datasets, for SVMs with the linear (first and third row) and the RBF (second and fourth row) kernel, trained with $C=1$ and $\gamma=0.5$.
The original data distribution is shown in the first column (first and second row for the linearly-separable data, third and fourth row for the parabolically-separable data). The decision boundaries of SVMs trained in the absence of label flips, and the corresponding test errors are shown in the second column.
The remaining columns report the results for each of the seven considered attack strategies, highlighting the corresponding $L=20$ label flips (out of 200 training samples).}
\vspace{-5pt}
\label{fig:exp_syn01}
\end{figure*}

To intuitively understand the fundamental strategies and differences of each of the proposed adversarial label flip attacks, we report here an experimental evaluation on two bi-dimensional datasets, where the positive and the negative samples can be perfectly separated by a linear and a parabolic decision boundary, respectively.\footnote{Data is available at \url{http://home.comcast.net/~tom.fawcett/public_html/ML-gallery/pages/index.html}.}
For these experiments, we learn SVMs with the linear and the RBF kernel on both datasets, using \texttt{LibSVM}~\cite{libsvm}. We set the regularization parameter $C=1$, and the kernel parameter $\gamma=0.5$, based on some preliminary experiments. For each dataset, we randomly select $200$ training samples, and evaluate the test error on a disjoint set of $800$ samples. The proposed attacks are used to flip $L=20$ labels in the training data (\ie, a fraction of 10\%), and the SVM model is subsequently learned on the tainted training set. 
Besides the four proposed attack strategies for adversarial label noise, further three attack strategies are evaluated for comparison, respectively referred to as \texttt{farfirst}, \texttt{nearest}, and \texttt{random}. As for \texttt{farfirst} and \texttt{nearest}, only the labels of the $L$  farthest and of the $L$ nearest samples to the decision boundary are respectively flipped. 
As for the \texttt{random} attack, $L$ training labels are randomly flipped. To mitigate the effect of randomization, each random attack selects the best label flips over $10$ repetitions.

Results are reported in Fig.~\ref{fig:exp_syn01}.
First, note how the proposed attack strategies \texttt{alfa}, \texttt{alfa-cr}, \texttt{alfa-tilt}, and \texttt{correlated cluster} generally exhibit clearer patterns of flipped labels than those shown by \texttt{farfirst},  \texttt{nearest}, and \texttt{random}, yielding indeed higher error rates.
In particular, when the RBF kernel is used, the SVM's performance is significantly affected by a careful selection of training label flips (\cf{} the error rates between the plots in the first and those in the second row of Fig.~\ref{fig:exp_syn01}). This somehow contradicts the result in~\cite{christmann04}, where the use of bounded kernels has been advocated to improve robustness of SVMs against training data perturbations. The reason is that, in this case, the attacker does not have the ability to make unconstrained modifications to the feature values of some training samples, but can only flip a maximum of $L$ labels. As a result, bounding the feature space through the use of bounded kernels to counter label flip attacks is not helpful here.
Furthermore, the security of SVMs may be even worsened by using a non-linear kernel, as it may be easier to significantly change (\eg, ``bend'') a non-linear decision boundary using carefully-crafted label flips, thus leading to higher error rates.
Amongst the attacks, \texttt{correlated cluster} shows the highest error rates when the linear (RBF) kernel is applied to the linearly-separable (parabolically-separable) data. In particular, when the RBF kernel is used on the parabolically-separable data, even only $10\%$ of label flips cause the test error to increase from $2.5\%$ to $21.75\%$. Note also that \texttt{alfa-cr} and \texttt{alfa-tilt} outperform \texttt{alfa} on the linearly-separable dataset, but not on the parabolically-separable data.
Finally, it is worth pointing out that applying the linear kernel on the parabolically-separable data, in this case, already leads to high error rates, making thus difficult for the label flip attacks to further increase the test error (\cf{} the plots in the third row of Fig.~\ref{fig:exp_syn01}).

\subsection{On Real-World Datasets}

\begin{figure*}[htb!]
\centering
\includegraphics[width=0.99\textwidth]{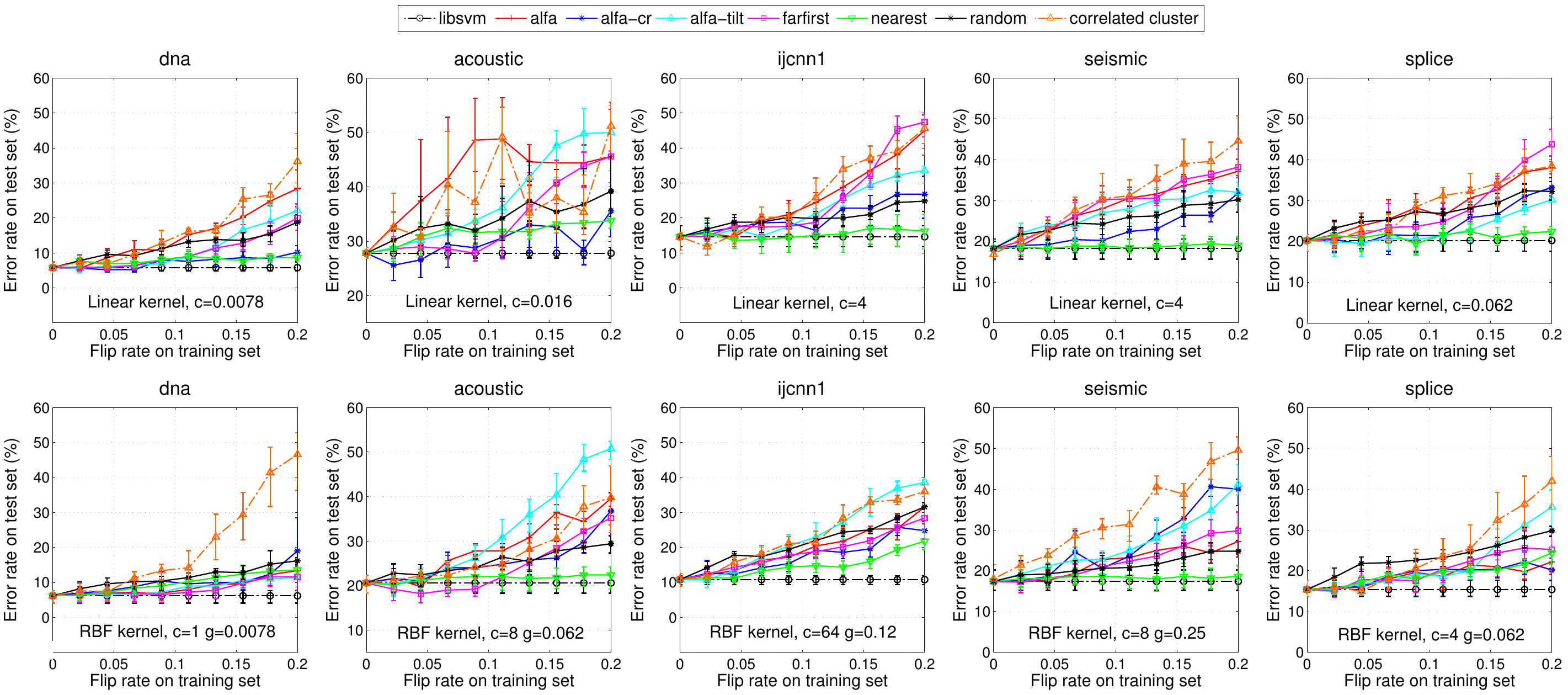}
\caption{Results on real-world datasets (in different columns), for SVMs with the linear (first row) and the RBF (second row) kernel. Each plot shows the average error rate ($\pm$ \emph{half} standard deviation, for readability) for each attack strategy,
estimated from $500$ samples using 5-fold cross validation, against an increasing fraction of adversarially flipped labels. The values of $C$ and $\gamma$ used to learn the SVMs are also reported for completeness (\cf{} Table~\ref{tbl:data}).}
\vspace{-5pt}
\label{fig:exp_real1a}
\end{figure*}

We report now a more systematic and quantitative assessment of our label flip attacks against SVMs, considering five real-world datasets publicly available at the \texttt{LibSVM} website.\footnote{\url{http://www.csie.ntu.edu.tw/~cjlin/libsvmtools/datasets/binary.html}}
In these experiments, we aim at evaluating how the performance of SVMs decreases against an increasing fraction of adversarially flipped training labels, for each of the proposed attacks. This will indeed allow us to assess their effectiveness as well as the security of SVMs against adversarial label noise.
For a fair comparison, we randomly selected $500$ samples from each dataset, 
and select the SVM parameters from $C \in \{ 2^{-7}, 2^{-6},\ldots,2^{10}\}$ and $\gamma \in \{ 2^{-7},2^{-6},\ldots,2^{5}\}$ by a $5$-fold cross validation procedure. 
The characteristics of the datasets used, along with the optimal values of $C$ and $\gamma$ as discussed above, are reported in Table~\ref{tbl:data}.
We then evaluate our attacks on a separate test set of $500$ samples, using 5-fold cross validation.
The corresponding average error rates are reported in Fig.~\ref{fig:exp_real1a}, 
against an increasing fraction of label flips, for each considered attack strategy, and for SVMs trained with the linear and the RBF kernel.

\begin{table}[t]
\small
\centering
\begin{tabular}{|c|c|c|}
\hline
\multicolumn{3}{|c|}{Characteristics of the real-world datasets} \\ \hline
Name & Feature set size & SVM parameters \\ \hline
\multirow{2}{*}{dna} & \multirow{2}{*}{$124$} & Linear kernel $C=0.0078$\\\cline{3-3}
& & RBF kernel $C=1, \gamma=0.0078$ \\\hline
\multirow{2}{*}{acoustic} & \multirow{2}{*}{$51$} & Linear kernel $C=0.016$\\\cline{3-3} 
& & RBF kernel $C=8, \gamma=0.062$ \\\hline
\multirow{2}{*}{ijcnn1} & \multirow{2}{*}{$23$} & Linear kernel $C=4$ \\\cline{3-3}
& & RBF kernel $C=64, \gamma=0.12$ \\\hline
\multirow{2}{*}{seismic} & \multirow{2}{*}{$51$} & Linear kernel $C=4$\\\cline{3-3} 
& & RBF kernel $C=8, \gamma=0.25$\\ \hline
\multirow{2}{*}{splice} & \multirow{2}{*}{$60$} & Linear kernel $C=0.062$\\\cline{3-3}
& & RBF kernel $C=4, \gamma=0.062$ \\\hline
\end{tabular}
\caption{Feature set sizes and SVM parameters for the real-world datasets.}
\vspace{-10pt}
\label{tbl:data}
\end{table}

The reported results show how the classification performance is degraded by the considered attacks, against an increasing percentage of adversarial label flips.
Among the considered attacks, \texttt{correlated cluster} shows an outstanding capability of subverting SVMs, although requiring significantly increased computational time.
In particular, this attack is able to induce a test error of almost $50\%$ on the \textit{dna} and \textit{seismic} data, when the RBF kernel is used. 
Nevertheless, \texttt{alfa} and \texttt{alfa-tilt} can achieve similar results on the \textit{acoustic} and \textit{ijcnn1} data, while being much more computationally efficient.
In general, all the proposed attacks show a similar behavior for both linear and RBF kernels, and lead to higher error rates on most of the considered real-world datasets than the trivial attack strategies \texttt{farfirst}, \texttt{nearest}, and \texttt{random}. For instance, when $20\%$ of the labels are flipped, \texttt{correlated cluster}, \texttt{alfa}, \texttt{alfa-cr}, and \texttt{alfa-tilt}  almost achieve an error rate of $50\%$, while \texttt{farfirst}, \texttt{nearest} and \texttt{random} hardly achieve an error rate of $30\%$. It is nevertheless worth remarking that \textit{farfirst} performs rather well against linear SVMs, while being not very effective when the RBF kernel is used. This reasonably means that non-linear SVMs may not be generally affected by label flips that are \emph{far} from the decision boundary. 

To summarize, our results demonstrate that SVMs can be significantly affected by the presence of well-crafted, adversarial label flips in the training data, which can thus be considered a relevant and practical security threat in application domains where attackers can tamper with the training data.

\section{Conclusions and Future Work}
\label{sect:conclusions}

Although (stochastic) label noise has been well studied especially in
the machine learning literature (\eg, see \cite{frenay14} for a
survey), to our knowledge few have investigated the
robustness of learning algorithms against well-crafted, malicious
label noise attacks.  In this work, we have focused on the problem of
learning with label noise from an adversarial perspective, extending
our previous work on the same topic~\cite{biggio11-acml,xiao12}.
In particular, we have discussed a framework that encompasses
different label noise attack strategies, revised our two previously-proposed label flip attacks accordingly, and presented two novel attack strategies that can significantly worsen the SVM's classification performance on untainted test data, even if only a small fraction of the training labels are manipulated by the attacker.  

An interesting future extension of this work may be to consider adversarial label noise attacks in which the attacker has limited knowledge of the system,
\eg, when the feature set or the training data are not completely
known to the attacker, to see whether the resulting error remains
considerable also in more practical attack scenarios.
Another limitation that may be easily overcome in the future is the assumption of equal cost for each label flip. In general, indeed, a different cost can be incurred depending on the feature values of the considered sample.

We nevertheless believe that this work provides an interesting starting point for future investigations on this topic, and may serve as a foundation for
designing and testing SVM-based learning algorithms to be more robust
against a deliberate label noise injection. To this end, inspiration
can be taken from previous work on robust SVMs to stochastic label
noise~\cite{stempfel09,biggio11-acml}.
Alternatively, one may exploit our framework to simulate a \emph{zero-sum game} between the attacker and the classifier, that respectively aim to maximize and  minimize the classification error on the untainted test set. This essentially amounts to re-training the classifier on the tainted data, \emph{having knowledge} of which labels might have been flipped, to learn a more secure classifier. Different game formulations can also be exploited if the players use non-antagonistic objective functions, as in \cite{bruckner12}.

We finally argue that our work may also provide useful insights for developing novel techniques in machine learning areas which are not strictly related to adversarial learning, such as semi-supervised and active learning. In the former case, 
by turning the maximization in Problems~\eqref{eq:attacker-obj}-\eqref{eq:svm-attack-obj} into a minimization problem,
one may find suitable label assignments $\vct z$ for the unlabeled data, thus effectively designing a semi-supervised learning algorithm.  Further, by exploiting the continuous label relaxation of Sect.~\ref{sect:alfa-cr}, one can naturally implement a \emph{fuzzy} approach, mitigating the influence of potentially outlying instances. 
As for active learning, minimizing the objective of Problems~\eqref{eq:attacker-obj}-\eqref{eq:svm-attack-obj} may help identifying the training labels which may have a higher impact on classifier training, \ie, some of the most informative ones to be queried. Finally, we conjecture that our approach may also be applied in the area of structured output prediction, in which semi-supervised and active learning can help solving the inference problem of finding the best structured output prediction approximately, when the computational complexity of that problem is not otherwise tractable.
 
\section*{Acknowledgments}
We would like to thank the anonymous reviewers for providing useful insights on how to improve our work. This work has been partly supported by the project ``Security of pattern recognition systems in future internet'' (CRP-18293) funded by Regione Autonoma della Sardegna, L.R. 7/2007, Bando 2009, and by the project ``Automotive, Railway and Avionics Multicore Systems - ARAMiS'' funded by the Federal Ministry of Education and Research of Germany, 2012.

\medskip
\parpic{\includegraphics[width=1in,clip,keepaspectratio]{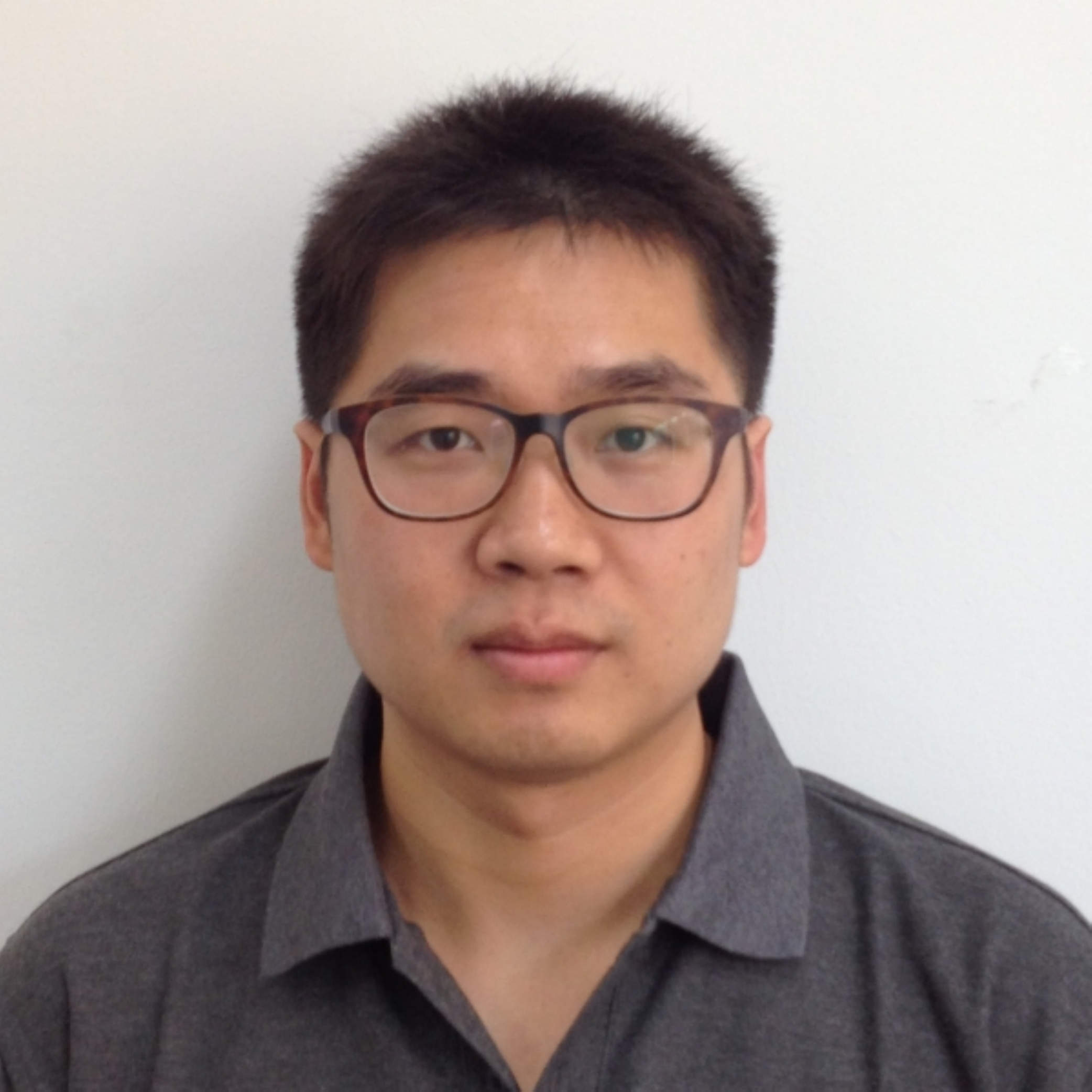}}
\noindent {\bf Huang Xiao} got his B.Sc. degree in Computer Science from the Tongji University in Shanghai in China. After that, he got the M.Sc. degree in Computer Science from the Technical University of Munich (TUM) in Germany. His research interests include semi-supervised and nonparametric learning,  machine learning in anomaly detection, adversarial learning, causal inference, Bayesian network, Copula theory. He is now a 3rd year PhD candidate at the Dept. of Information Security at TUM, supervised by Prof. Dr. Claudia Eckert.

\medskip
\parpic{\includegraphics[width=1in,clip,keepaspectratio]{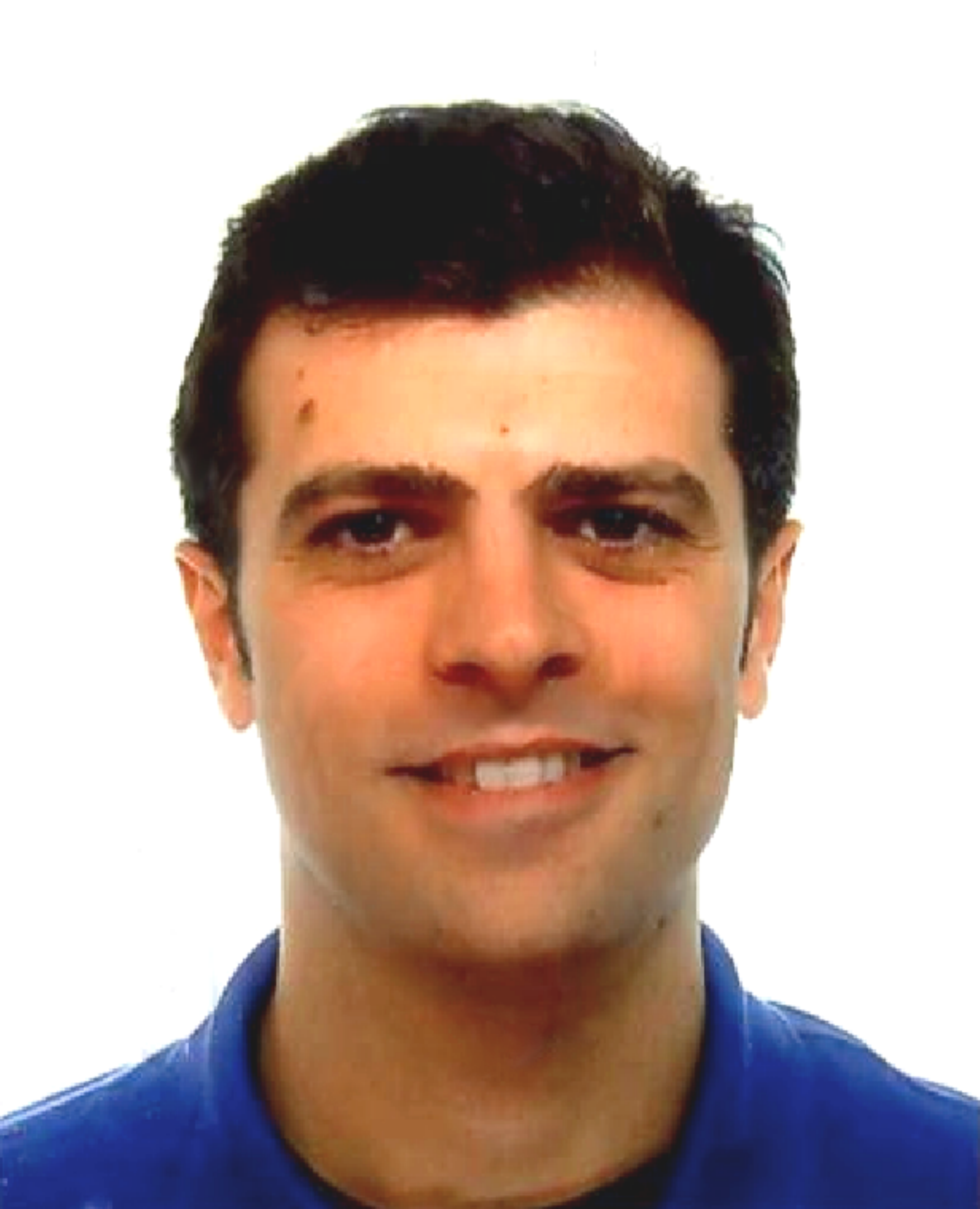}}
\noindent {\bf Battista Biggio} received the M. Sc. degree in Electronic Eng., with honors, and the Ph. D. in Electronic Eng. and Computer Science, respectively in 2006 and 2010, from the University of Cagliari, Italy.
Since 2007 he has been working for the Dept. of Electrical and Electronic Eng. of the same University, where he is now a postdoctoral researcher. 
In 2011, he visited the University of T\"ubingen, Germany, for six months, and worked on the security of machine learning algorithms to training data contamination.
His research interests currently include: secure / robust machine learning and pattern recognition methods, multiple classifier systems, kernel methods, biometric authentication, spam filtering, and computer security.
He serves as a reviewer for several international conferences and journals, including Pattern Recognition and Pattern Recognition Letters.
Dr. Biggio is a member of the IEEE Computer Society and IEEE Systems, Man and Cybernetics Society, and of the Italian Group of Italian Researchers in Pattern Recognition (GIRPR), affiliated to the International Association for Pattern Recognition.

\medskip
\parpic{\includegraphics[width=1in,clip,keepaspectratio]{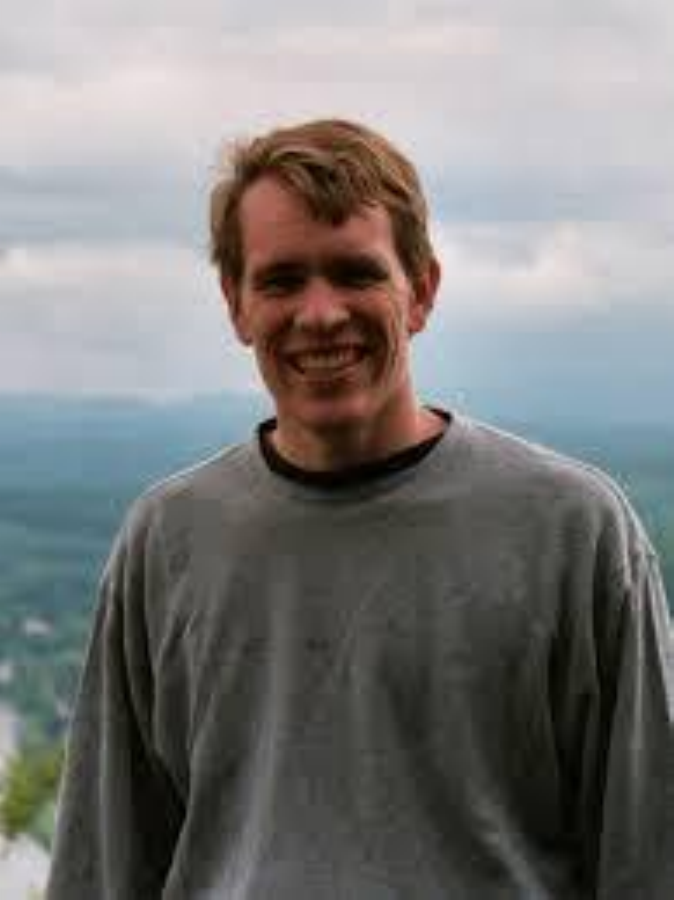}}
\noindent {\bf Blaine Nelson} is currently a postdoctoral researcher at the University of Cagliari, Italy. He previously was a postdoctoral research fellow at the University of Potsdam and at the University of T\"ubingen, Germany, and completed his doctoral studies at the University of California, Berkeley. Blaine was a co-chair of the 2012 and 2013 AISec workshops on artificial intelligence and security and was a co-organizer of the Dagstuhl Workshop ``Machine Learning Methods for Computer Security'' in 2012. His research focuses on learning algorithms particularly in the context of security-sensitive application domains. Dr. Nelson investigates the vulnerability of learning to security threats and how to mitigate them with resilient learning techniques.

\medskip
\parpic{\includegraphics[width=1in,clip,keepaspectratio]{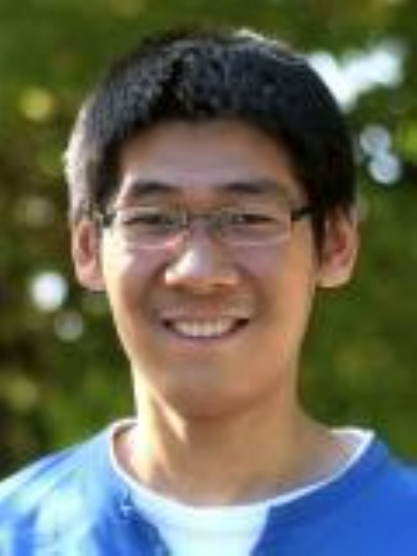}}
\noindent {\bf Han Xiao} is a Ph.D. candidate at Technical University of Munich (TUM), Germany. He got the M.Sc. degree at TUM in 2011. His advisors are Claudia Eckert and Ren\'e Brandenberg. His research interests include online learning, semi-supervised learning, active learning, Gaussian process, support vector machines and probabilistic graphical models, as well as their applications in knowledge discovery. He is supervised by Claudia Eckert and Ren\'e Brandenberg. From Sept. 2013 to Jan. 2014, he was a visiting scholar in Shou-De Lin's Machine Discovery and Social Network Mining Lab at National Taiwan University.

\medskip
\parpic{\includegraphics[width=1in,clip,keepaspectratio]{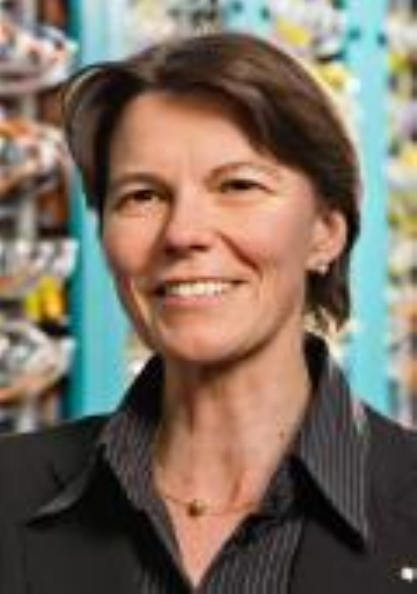}}
\noindent {\bf Claudia Eckert} got her diploma in computer science from the University of Bonn. And she got the PhD in 1993 and in 1999 she completed her habilitation at the TU Munich on the topic ``Security in Distributed Systems''. Her research and teaching activities can be found in the fields of operating systems, middleware, communication networks and information security. In 2008, she founded the center in Darmstadt CASED (Center for Advanced Security Research Darmstadt), she was the deputy director until 2010. She is a member of several scientific advisory boards, including the Board of the German Research Network (DFN), OFFIS, Bitkom and the scientific committee of the Einstein Foundation Berlin. She also advises government departments and the public sector at national and international levels in the development of research strategies and the implementation of security concepts. Since 2013, she is the member of the bavarian academy of science.

\medskip
\parpic{\includegraphics[width=1in,clip,keepaspectratio]{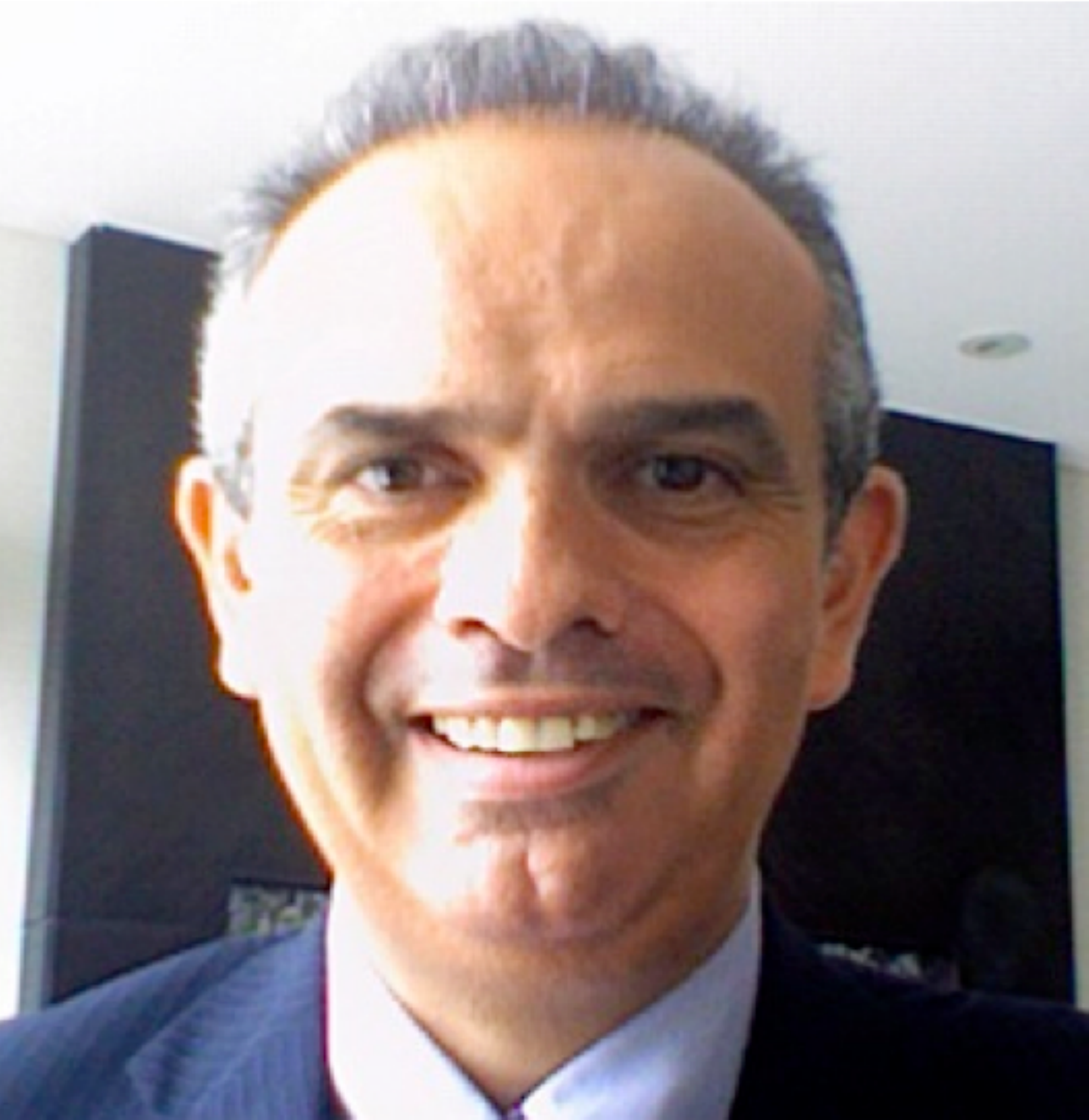}}
\noindent {\bf Fabio Roli} received his M. Sc. degree, with honors, and Ph. D. degree in Electronic Eng. from the University of Genoa, Italy. He was a member of the research group on Image Processing and Understanding of the University of Genoa, Italy, from 1988 to 1994. He was adjunct professor at the University of Trento, Italy, in 1993 and 1994. In 1995, he joined the Dept. of Electrical and Electronic Eng. of the University of Cagliari, Italy, where he is now professor of computer engineering and head of the research group on pattern recognition and applications. His research activity is focused on the design of pattern recognition systems and their applications to biometric personal identification, multimedia text categorization, and computer security. On these topics, he has published more than two hundred papers at conferences and on journals. He was a very active organizer of international conferences and workshops, and established the popular workshop series on multiple classifier systems. Dr. Roli is a member of the governing boards of the International Association for Pattern Recognition and of the IEEE Systems, Man and Cybernetics Society. He is Fellow of the IEEE, and Fellow of the International Association for Pattern Recognition.

\end{document}